\documentclass{article}
\usepackage{graphicx} %
\usepackage[super, comma, sort&compress]{natbib}
\usepackage{booktabs}
\usepackage{xcolor}
\usepackage[most]{tcolorbox}
\usepackage{multirow}
\usepackage{multicol}

\newtcbox{\code}{
  on line, boxrule=0pt, boxsep=0pt,
  left=2pt, right=2pt, top=1pt, bottom=1pt,
  arc=2pt, colback=gray!25, colframe=gray!25,,
  fontupper=\ttfamily\small
}

\definecolor{linkcol}{HTML}{3073AD}  %
\definecolor{citecol}{HTML}{3073AD} %
\definecolor{urlcol}{HTML}{3073AD} %

\definecolor{lightblue}{RGB}{32, 194, 217}
\definecolor{jku_red}{RGB}{217, 92, 76}
\definecolor{jku_blue}{RGB}{0, 132, 187}
\definecolor{jku_green}{RGB}{91, 167, 85} %
\definecolor{jku_yellow}{RGB}{241, 188, 63}
\definecolor{jku_cyan}{RGB}{79,176,191}
\definecolor{jku_grey}{RGB}{125,130,140}
\definecolor{jku_lightgreen}{RGB}{191,206,82}
\definecolor{jku_violett}{RGB}{174,97,157}

\usepackage{hyperref}
\hypersetup{
   colorlinks=true,
   linkcolor=linkcol,
   citecolor=citecol,
   urlcolor=urlcol,
}
\usepackage{caption}
\usepackage[preprint]{neurips_2026}%

\title{BatchBench: Benchmarking AI Robustness to Batch Effects and Domain Shift in Biomedical Images \\ 
-- A JUMP Cell Painting Leaderboard}

\author{%
    \textbf{Sonja Topf}$^{1,2}$ \quad
    \textbf{Ana Sanchez-Fernandez}$^{1}$ \quad
    \textbf{Günter Klambauer}$^{1,2}$\\
    $^{1}$ELLIS Unit Linz and LIT AI Lab, Institute for Machine Learning,\\
    $^{2}$Clinical Research Institute for Medical AI\\
    Johannes Kepler University, Linz, Austria\\
}

\begin{document}

\maketitle
\vspace{-20pt}

\begin{abstract} \bf
    Despite recent advances, current AI systems remain highly susceptible to domain shifts, particularly in biomedical imaging, where images are usually acquired under substantially different experimental conditions. Although recent domain adaptation methods have made substantial progress, significant performance gaps under domain shift persist. To measure progress in this regard, we introduce \textit{BatchBench}, a leaderboard that evaluates and compares the robustness of AI systems on biomedical images from unseen experimental batches and data sources. \textit{BatchBench} is built on JUMP-CP, one of the largest publicly available biomedical imaging datasets. Generated using the Cell Painting assay, JUMP-CP exhibits pronounced batch effects and domain shifts, making it a challenging testbed for evaluating robustness to distributional changes. The leaderboard is open for submissions on \url{https://huggingface.co/spaces/ai-lab-jku/jumpcp-challenge}.

    Keywords: biomedical imaging, domain-adaptation, test-time-adaptation, cell-painting, jump-cp, microscopy, distribution-shift
\end{abstract}

\section{Introduction}
\begin{figure}[hb]
    \centering
    \includegraphics[width=0.9\linewidth]{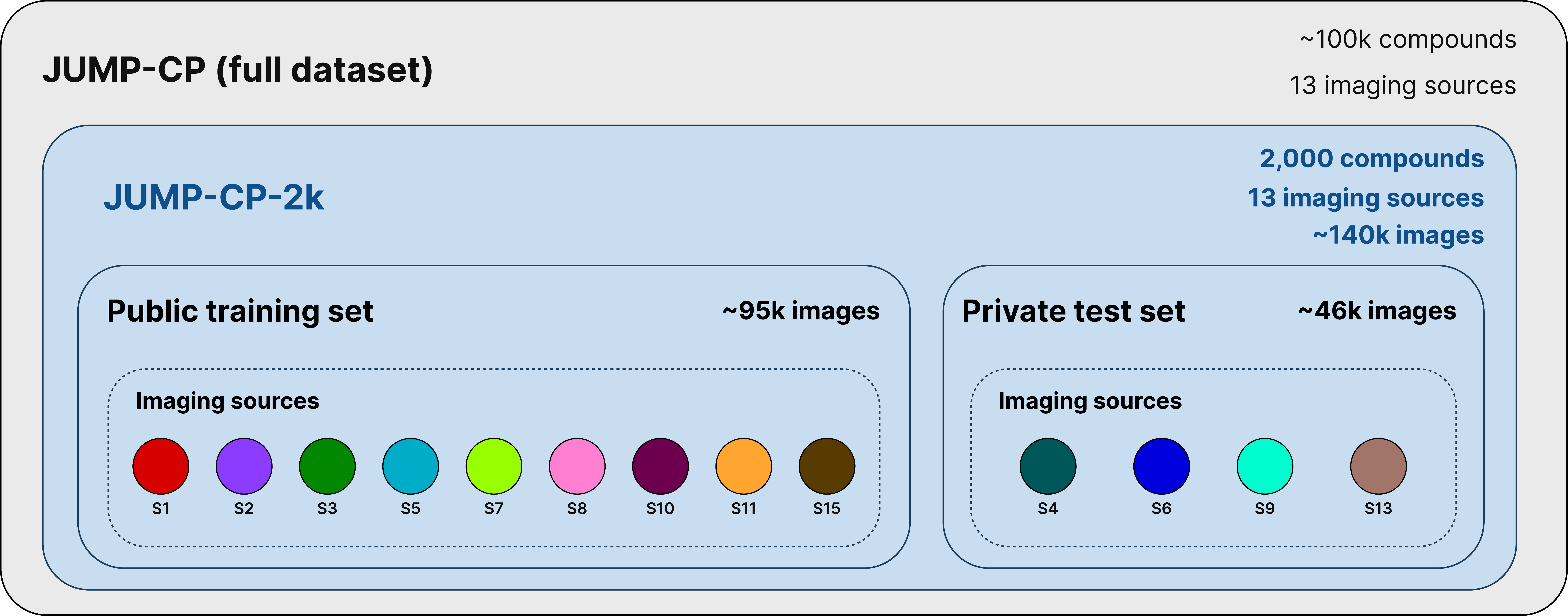}
    \caption{Source-level split of the BatchBench leaderboard. Nine
    sources of JUMP-CP-2k are public (eight for training, source~2 for
    validation). Sources 4, 6, 9, and 13 are held out as the private
    test set, so that every leaderboard entry is evaluated on unseen
    data-generating sources.}
    \label{fig:dataset_overview}
\end{figure}

\paragraph{Vulnerability of AI systems under domain shift and batch effects.} Deep learning methods often assume that training and test data are sampled from the same underlying distribution.  However, in practice, biological images are acquired across experimental batches, plates, and sites, which introduces non-biological variation known as \emph{batch effects}. These effects shift the distribution of the input images, such that a model that performs well on its training distribution can fail on images from a new experimental batch or laboratory. This limits
the use of AI in applications such as drug discovery, where new data
are generated continuously and under changing conditions.

\paragraph{JUMP-CP and CellPainting.} High-content screening data obtained with the Cell Painting assay~\cite{Bray2016} provide a clear example of this problem. The JUMP Cell Painting dataset (JUMP-CP)~\cite{JUMP-CP}, one of the largest public biomedical imaging datasets, was obtained from 13 data-generating centers (\emph{sources}) using different microscopes and laboratory workflows. As a result, JUMP-CP exhibits substantial batch effects across multiple levels of its data hierarchy. Because every image is annotated with its source, experimental batch, and plate, JUMP-CP provides a exceptional setting in which domain shifts are both pronounced and explicitly annotated at multiple levels.

\paragraph{Domain adaptation (DA)} methods address domain shift either by learning
domain-invariant representations during training or by adapting the
model to unlabeled target data at inference time, i.e. test-time adaptation (TTA) \cite{tent, deepCORAL, Dann, AdaBN, ARM-BN, sanchezfernandez2026stabilizing}. These methods have substantially improved robustness on natural
images. Nevertheless, their effectiveness on biomedical images from unseen experimental conditionsremains poorly characterized, and results reported on different subsets, splits, and metrics are rarely
comparable.

\paragraph{The BatchBench Leaderboard.}
We introduce the \emph{BatchBench leaderboard}, an open benchmark for comparing AI models on the identification of chemical perturbations in Cell Painting images acquired from sources not seen during training. The leaderboard is designed to be used by several research communities:

\begin{itemize}
\item \textbf{Domain adaptation and test-time adaptation.} \textit{BatchBench} evaluates multi-source adaptation to unseen target domains under combined covariate and label shift. Methods that adapt to unlabeled test images, as well as methods that do not perform test-time adaptation, can be submitted through a common \code{adapt()}/\code{predict()} interface.

\item \textbf{Biomedical imaging and foundation models.} \textit{BatchBench} evaluates whether pretrained image encoders and specialized architectures generalize to unseen imaging sites, an important requirement for deployment that is not captured by in-distribution evaluation.

\item \textbf{Cell Painting and JUMP-CP.}
\textit{BatchBench} provides a standardized source-level evaluation of image-based compound profiling on JUMP-CP using fixed splits and a private test set.

\end{itemize}

The leaderboard presents three components: (i)~\textbf{JUMP-CP-2k}, a
curated subset of JUMP-CP covering 2,000 compounds, with nine public
sources and four sources held out as a private test set
(Section~\ref{sec:leaderboard}); (ii)~\textbf{a containerized
submission protocol} that keeps the test data private and supports
both conventional and test-time adaptation methods
(Section~\ref{sec:protocol}); and (iii)~\textbf{eight baseline entries}
covering standard supervised training, domain-invariant training,
test-time adaptation, meta-learning, and a zero-shot multimodal LLM
(Sections~\ref{sec:baselines}
and~\ref{sec:results}). The baseline entries show that test-time
adaptation reduces the domain gap but does not close it. For compounds
with few training images, every current entry remains close to chance
level. The leaderboard is open for submissions on
\href{https://huggingface.co/spaces/ai-lab-jku/jumpcp-challenge}{Hugging
Face}.

\section{The BatchBench leaderboard}
\label{sec:leaderboard}

\paragraph{Problem setting.}
Let $\mathcal{D}^{(1)}, \ldots, \mathcal{D}^{(S)}$ denote labeled
data from $S$ training sources, where each sample consists of an
image $\boldsymbol{x}$, its compound label $y \in \{1, \ldots, K\}$,
and its domain metadata $\boldsymbol{d} = (\text{source},
\text{batch}, \text{plate})$. At test time, a submission receives the
unlabeled images $\{\boldsymbol{x}_n\}_{n=1}^{N}$ and their metadata
$\{\boldsymbol{d}_n\}_{n=1}^{N}$ from sources that do not occur in the
training data. It may adapt to these images before predicting their
labels. The leaderboard thus evaluates \emph{multi-source domain
adaptation to unseen target domains}. The setting covers domain
generalization (no adaptation), test-time adaptation, and in-context
adaptation as special cases. As the the domain metadata are
hierarchical, submissions can define domains at the level of
sources, experimental batches, or plates.

\paragraph{Task.}
Each input is a single five-channel Cell Painting image (DNA, RNA, ER,
AGP, and Mito). The task is to identify which of the $K = 2{,}000$
compounds the imaged cells were treated with. Test images come
exclusively from held-out sources, so every test image also originates
from an unseen experimental batch and an unseen plate.

\paragraph{Compound selection and sampling.}
JUMP-CP comprises images from 13 sources and more than 100,000 compound
and genetic perturbations~\cite{JUMP-CP}. We used the dataset cpg0016 \citep{JUMP-CP}, available from the Cell Painting Gallery \citep{Weisbart2024-ta} on the Registry of Open Data on AWS (https://registry.opendata.aws/cellpainting-gallery/). To obtain a dataset that is
practical to download and train on, we selected the 2,000 most frequent
compounds, retained all wells treated with these compounds, and sampled
one image (field of view) per well uniformly at random.

\paragraph{Domain metadata and controls.}
Each sample is annotated with its source, experimental batch, and plate
identifiers. Submissions may use these identifiers as domain labels
during training and to group test images at inference. Images of
negative-control wells (DMSO-treated, unperturbed cells) from the
training sources are provided as a separate file. They can be used,
for example, for control-based normalization \cite{sanchezfernandez2026stabilizing}.

\paragraph{Preprocessing.}
From each original image we extract a $500\times500$ pixel center crop.
We then apply illumination correction by clipping the top 0.01\% of
pixel intensities in each channel.

\paragraph{Splits.}
Sources 4, 6, 9, and 13 form the private test set of the leaderboard
and are never released (Figure~\ref{fig:dataset_overview}). The
remaining nine sources are public. Source~2 serves as the validation
set for model selection, and the other eight sources form the training
set. 

\paragraph{Evaluation subsets.}
Compounds differ greatly in their number of training images and in the
strength of their phenotypes. In addition to the overall accuracy, the
leaderboard therefore reports accuracy on three subsets of
the test set:
\begin{itemize}
    \item \texttt{POSCON}: the eight positive-control compounds, present
    on every plate, usually at fixed well positions. They have many
    training images and strong,
    well-characterized phenotypes.
    \item \texttt{TARGET}: 293 compounds from the JUMP-Target
    collection, a reference library chosen to cover diverse biological
    targets and mechanisms of action.
    \item \texttt{REST}: all remaining compounds, each represented by
    few images (about 12 per compound in the public split).
\end{itemize}
The three subsets correspond to increasingly difficult regimes:
abundant data with strong phenotypes (\texttt{POSCON}), a curated
mechanism-of-action reference set (\texttt{TARGET}), and a long tail
of rare classes (\texttt{REST}).

\paragraph{Label shift.}
In addition to the source shift, the class distribution differs
between the public and the test split. \texttt{POSCON} images make up
51\% of the public split but 36\% of the test set; \texttt{TARGET}
images increase from 25\% to 55\%; \texttt{REST} images decrease from
24\% to 9\% (Table~\ref{tab:dataset-stats}). The leaderboard therefore
evaluates adaptation under combined covariate and label shift, a
setting in which many TTA methods are known to
degrade~\cite{Park2023, Zhao2023}.

\begin{table}[h]
\centering
\small
\setlength{\tabcolsep}{4pt}
\caption{Statistics of JUMP-CP-2k per split and evaluation subset. The
private test set of the leaderboard consists of the held-out sources
4, 6, 9, and 13. Samples/compound is the mean number of images per
compound. Source, batch, and plate identifiers are provided for every
sample.}
\label{tab:dataset-stats}
\begin{tabular}{@{}llrrrrrr@{}}
\toprule
Split & Subset & Compounds & Samples & Samples/compound & Sources & Batches & Plates \\
\midrule
\multirow{3}{*}{Public (train + val)}
 & \texttt{POSCON} & 8 & 48,574 & 6,071.75 & 9 & 102 & 1,534 \\
 & \texttt{TARGET} & 293 & 23,455 & 80.05 & 7 & 63 & 76 \\
 & \texttt{REST} & 1,926 & 23,186 & 12.04 & 9 & 85 & 931 \\
\midrule
\multirow{3}{*}{Private test}
 & \texttt{POSCON} & 8 & 16,691 & 2,086.38 & 4 & 42 & 370 \\
 & \texttt{TARGET} & 293 & 25,352 & 86.53 & 4 & 40 & 62 \\
 & \texttt{REST} & 1,864 & 4,107 & 2.20 & 2 & 18 & 179 \\
\bottomrule
\end{tabular}
\end{table}

\paragraph{Research questions.}
The leaderboard is designed to answer questions that are open in each
of the communities above, for example:
\begin{itemize}
    \item Do domain adaptation and TTA methods developed on natural-image
    benchmarks transfer to biological domain shift?
    \item At which level of the domain hierarchy (source, batch, or
    plate) should adaptation be performed?
    \item Do pretrained microscopy foundation models generalize to unseen
    imaging sources better than models trained from scratch?
    \item Can models identify compounds with weak phenotypes and few
    training images (\texttt{TARGET}, \texttt{REST}) on unseen sources?
\end{itemize}

\section{Submitting to the leaderboard}
\label{sec:protocol}

\paragraph{Submission format.}
A leaderboard submission is a public Git repository together with a
commit hash. The repository must contain the training code and a
Dockerfile from which the evaluation server builds a container image.
Submissions are therefore independent of the deep learning framework
and model architecture. The image must provide a
\code{Submission} class that implements two methods, \code{adapt()} and
\code{predict()}. Ready-to-run example submissions for all baseline
entries are available in our
\href{https://github.com/ml-jku/batchbench-baselines/}{baseline
repository} and can serve as templates.

\paragraph{Evaluation.}
The evaluation server passes the private test images, as an array of
shape $(N, H, W, 5)$, together with a metadata list containing the
source, batch, and plate identifier of every image, to both
\code{adapt()} and \code{predict()}. No labels are provided.
\code{adapt()} may update the model using the unlabeled test images
and their metadata, which permits test-time adaptation. Methods that
do not adapt should implement it as a no-op. \code{predict()} must return one
prediction per image, in the original order.

\paragraph{Metric and ranking.}
Leaderboard entries are ranked by overall accuracy, i.e., the fraction
of correctly classified images in the entire test set. Because
\texttt{POSCON} and \texttt{TARGET} together contain over 90\% of the
test images, this metric is dominated by these two subsets. The
leaderboard therefore additionally reports accuracy on
\texttt{POSCON}, \texttt{TARGET}, and \texttt{REST} separately, so
that progress on rare compounds remains visible. Scores are published
on the leaderboard once the manual evaluation finishes.

\paragraph{Rules.}
Submissions may not use any data from the held-out sources (4, 6, 9,
and 13) for training, pre-training, or model selection. This also
applies to pretrained models whose pre-training data include these
sources. Test-time adaptation on the unlabeled test images through
\code{adapt()} is permitted.

\section{Development datasets}

To lower the entry barrier and enable fast prototyping before a
leaderboard submission, we release two smaller variants of
JUMP-CP-2k:
\begin{itemize}
    \item \textbf{JUMP-CP-MNIST} contains only the eight
    positive-control compounds. Its images are downsampled to
    $200\times200$ pixels, and the five imaging channels (DNA, RNA, ER,
    AGP, and Mito) are combined into three channels, so that standard
    RGB architectures and pretrained models can be applied without
    modification.
    \item \textbf{JUMP-CP-2k-200} contains all 2,000 compounds of
    JUMP-CP-2k at the reduced resolution of $200\times200$ pixels, with
    all five channels.
\end{itemize}
Both variants use the same source-level split as JUMP-CP-2k.
Leaderboard rankings are computed on the full-resolution private
test set.

\section{Baseline entries}
\label{sec:baselines}

The leaderboard is initialized with eight baseline entries that
represent the main strategies for handling domain shift. All except the
LLM entry use a ResNet18~\cite{ResNet18} backbone whose first
convolution takes five input channels.

\begin{itemize}
    \item \textbf{No adaptation (ResNet18).} A classifier trained with
    standard empirical risk minimization on the training sources and
    applied to the test sources unchanged.
    \item \textbf{Domain-invariant training: DANN~\cite{Dann} and
    deepCORAL~\cite{deepCORAL}.} Both methods encourage representations
    that do not depend on the domain during training, adversarially
    (DANN) or by aligning second-order feature statistics (deepCORAL).
    We use the source identifier as the domain label.
    \item \textbf{Test-time adaptation: AdaBN~\cite{AdaBN} and
    TENT~\cite{tent}.} Both use the trained ResNet18 baseline and adapt
    it only at inference. AdaBN replaces the BatchNorm statistics with
    statistics computed on the test images. TENT additionally updates
    the BatchNorm affine parameters by minimizing the entropy of the
    predictions. For both methods, test images are grouped by plate, so
    that each plate is adapted separately.
    \item \textbf{Meta-learning: ARM-BN~\cite{ARM-BN}. and CS-ARM-BN \cite{sanchezfernandez2026stabilizing}} ARM-BN trains
    the model episodically to classify images using BatchNorm
    statistics computed from an unlabeled context batch of the same
    domain. At test time, the statistics are computed in the same way
    from the unlabeled test images.
    \item \textbf{Zero-shot multimodal LLM: Claude Opus 5.5.} Without
    any training, Claude Opus 5.5 (effort set to \code{high} ) is shown each test image as an RGB composite (red = AGP, green = ER + Mito, blue = DNA + RNA) together with all
    2,000 candidate compounds as SMILES strings, and is asked which
    compound induced the phenotype. Ten images share one request, and
    the model answers with one candidate number per image. Whether its
    pre-training data include JUMP-CP images is unknown.
\end{itemize}

Hyperparameters of the ResNet18-based entries were tuned with Bayesian
optimization, and for every method the model with the highest accuracy
on source~2 was selected. The code of these entries is available in our
\href{https://github.com/ml-jku/batchbench-baselines/}{GitHub repository}.
The LLM entry requires API access and therefore does not follow the
containerized submission protocol. We evaluated it on the private test
set and provide the code and raw model outputs
\href{https://github.com/ml-jku/batchbench-llm-baselines}{
here}.

\section{Results}
\label{sec:results}

Table~\ref{tab:results} reports the current leaderboard entries.

\paragraph{Test-time adaptation reduces the domain gap.}
Without adaptation, ResNet18 reaches an average overall accuracy of 0.175.
Re-estimating the BatchNorm statistics on each test plate (AdaBN)
raises this to 0.244. Most of this gain comes from \texttt{POSCON},
where accuracy increases from 0.465 to 0.646. TENT's additional
entropy minimization yields a higher performance in the \texttt{POSCON} group, but no improvement in \texttt{TARGET} or   \texttt{REST}. Domain-invariant training does not improve over the
non-adapted baseline: DANN reaches 0.170.

\paragraph{The domain gap remains for compounds with few images.}
Performance on \texttt{POSCON} is substantially higher than on \texttt{TARGET} and \texttt{REST}, consistent with the much larger number of training images per compound, lower number of classes and strong phenotypes. These factors are absent for most \texttt{TARGET} and \texttt{REST} compounds, and none of the current entries compensates for them. Identifying compounds with weak phenotypes and few training images on unseen sources is therefore the main open problem posed by the leaderboard.

\paragraph{A general-purpose LLM performs at chance level.}
Zero-shot, Claude Opus 5.5 identifies 41 of 46,150 test images
correctly (accuracy 0.0009, chance 0.0005), and none of its subset
accuracies is clearly above chance. A general-purpose multimodal LLM
therefore cannot match Cell Painting phenotypes to compound structures
without task-specific training.

\begin{table}[h]
\centering
\caption{Baseline entries of the BatchBench leaderboard, evaluated on
the private test set (held-out sources 4, 6, 9, and 13). Values are
the mean $\pm$ sample standard deviation over three runs with
different random seeds; the zero-shot Claude Opus 5.5 entry is a single
run. Chance level is $1/2{,}000 = 0.0005$.}
\label{tab:results}
\begin{tabular}{lc|ccc}
\toprule
\textbf{Entry} & \textbf{Overall} & \texttt{POSCON} & \texttt{TARGET} & \texttt{REST} \\
\midrule
ResNet18 (no adaptation) & $0.175 \pm 0.015$ & $0.465 \pm 0.036$ & $0.012 \pm 0.004$ & $0.002 \pm 0.000$ \\
DANN                    & $0.170 \pm 0.011$ & $0.465 \pm 0.026$ & $0.003 \pm 0.004$ & $0.001 \pm 0.001$ \\
deepCORAL               & $0.201 \pm 0.006$ & $0.502 \pm 0.024$ & $0.034 \pm 0.006$ & $0.007 \pm 0.002$ \\
AdaBN                   & $0.244 \pm 0.005$ & $0.646 \pm 0.005$ & $0.019 \pm 0.005$ & $0.002 \pm 0.001$ \\
TENT                    & $0.249 \pm 0.007$ & $0.661 \pm 0.011$ & $0.017 \pm 0.005$ & $0.002 \pm 0.001$ \\
CS-ARM-BN               & $0.256 \pm 0.004$ & $0.660 \pm 0.005$ & $0.031 \pm 0.004$ & $0.004 \pm 0.001$ \\
ARM-BN                  & $0.260 \pm 0.002$ & $0.665 \pm 0.003$ & $0.035 \pm 0.002$ & $0.004 \pm 0.001$ \\
\midrule
Claude Opus 5.5 (zero-shot) & $0.0009$ & $0.0001$ & $0.0015$ & $0.0005$ \\
\bottomrule
\end{tabular}
\end{table}

\section*{Acknowledgements}
\label{sec:acknowledgements}
We thank Anne Carpenter, Shantanu Singh, and Runxi Shen (Broad Institute of
MIT and Harvard) for helpful discussions during the design of the
leaderboard. 

\bibliography{bib}
\bibliographystyle{apalike}

\appendix

\end{document}